\documentclass[letterpaper, 10pt, conference]{ieeeconf}      % Use this line for a4 paper

\IEEEoverridecommandlockouts                              % This command is only needed if 
\usepackage[left=54pt,right=54pt,top=54pt,bottom=72pt]{geometry}
\usepackage{multirow}
\usepackage{hyperref} 
\usepackage{booktabs}
\usepackage{subcaption}
\usepackage{graphics} % for pdf, bitmapped graphics files
\usepackage{epsfig} % for postscript graphics files
\usepackage{times} % assumes new font selection scheme installed
\usepackage{amsmath} % assumes amsmath package installed
\usepackage{amssymb}
\usepackage{url}

\title{\LARGE \bf
Sim-to-Real Domain Adaptation for Deformation Classification}

\author{Joel Sol$^{1}$,  Jamil Fayyad$^{1}$,  Shadi Alijani$^{1}$ and Homayoun Najjaran$^{1}$ % <-this % stops a space
\thanks{$^{1}$ Faculty of Electrical and Computer Engineering, University of Victoria, Victoria, BC V8P 5C2, Canada
        {\tt\small \{joelsol, jfayyad, shadialijani, najjaran\}@uvic.ca}}%
\thanks{The authors have provided supplementary material (code) available at \protect\url{https://github.com/JoelESol/CASNet}}
}

\begin{document}

\maketitle
\thispagestyle{empty}
\pagestyle{empty}

%%%%%%%%%%%%%%%%%%%%%%%%%%%%%%%%%%%%%%%%%%%%%%%%%%%%%%%%%%%%%%%%%%%%%%%%%%%%%%%%
\begin{abstract}

Deformation detection is vital for enabling accurate assessment and prediction of structural changes in materials, ensuring timely and effective interventions to maintain safety and integrity. Automating deformation detection through computer vision is crucial for efficient monitoring, but it faces significant challenges in creating a comprehensive dataset of both deformed and non-deformed objects, which can be difficult to obtain in many scenarios. In this paper, we introduce a novel framework for generating controlled synthetic data that simulates deformed objects. This approach allows for the realistic modeling of object deformations under various conditions. Our framework integrates an intelligent adapter network that facilitates sim-to-real domain adaptation, enhancing classification results requiring limited real data from deformed objects. We conduct experiments on domain adaptation and classification tasks and demonstrate that our framework improves sim-to-real classification results compared to the simulation baseline. Our code is available \href{https://github.com/JoelESol/CASNet}{here}.
\end{abstract}

%%%%%%%%%%%%%%%%%%%%%%%%%%%%%%%%%%%%%%%%%%%%%%%%%%%%%%%%%%%%%%%%%%%%%%%%%%%%%%%%
\section{Introduction}

Deformation classification is a pivotal process in various engineering disciplines \cite{inaudi1994low, tong1997detection}, aimed at identifying changes in the shape or structure of materials or objects under various conditions. Traditionally, this task has been approached through methods that primarily include mechanical and manual inspection techniques \cite{dulieu2005deformation}, as well as the use of basic photographic and sensor-based data \cite{karrock2015pressure}. These conventional methods rely on the expertise of human inspectors, the application of mechanical gauges, and other instrumentation techniques to detect and categorize deformations. while these solutions are foundational, they come with several disadvantages that can impact their effectiveness and efficiency \cite{davtalab2022automated}. Traditional methods are prone to human errors as they often rely on human inspectors to evaluate and classify deformations. This can introduce subjectivity and variability in the results due to differences in experience, expertise, and judgments. Moreover, Scaling traditional methods to larger or more complex structures can be problematic. Manual methods are not only slower but also often impractical for very large or intricate designs without substantial investment in manpower and equipment.  

Deformation classification using computer vision represents a significant advancement over traditional methods \cite{zhuang2022review}. These methods harness the power of digital image processing and machine learning algorithms to detect and categorize structural changes with high precision and automation. The modern approach involves capturing images or video data of the material or structure, which are then analyzed using sophisticated algorithms that can identify changes in shape or texture. These algorithms can be trained to recognize specific patterns of deformation, enabling them to differentiate between normal variations and potential failures. The application of computer vision not only increases the accuracy and reduces the subjectivity associated with manual inspections but also dramatically speeds up the assessment process. 

One significant shortcoming of using computer vision for deformation classification is the substantial need for large and diverse datasets to effectively train the algorithms. These datasets must encompass a wide range of examples of both defective and non-defected materials to ensure the models can accurately differentiate and recognize various types of deformations. Acquiring such datasets often involves considerable time, expense, and logistical effort, as it requires the collection of comprehensive imagery under multiple conditions and stages of deformation. In this paper, we propose a Content-Aware Sim-to-Real Network (CASNet).

\section{Related Works}
The detection of deformations has been a critical area of research within the field of computer vision, essential for applications ranging from structural health monitoring \cite{dong2021review} to manufacturing and production quality control \cite{saini2014identification}. Over time, deformation detection approaches have significantly developed to improve the quality and capability of the proposed methods, leading to enhanced precision, reliability, and speed in identifying structural deformities. Existing approaches can be divided into two main streams: Classical approaches, characterized by their reliance on traditional image processing techniques and heuristic algorithms, and Deep Learning-based approaches. Classical approaches often involve edge detection, feature extraction, and template matching, leveraging geometric and photometric invariants to identify deformations. Erkan et al. propose the integration of gravimetric and magnetic fields through deformation analysis for near-surface detection problems \cite{erkan2012fusion}. 
In \cite{mares2011measurement}, Mares et al. use colour encoding to measure transient 3D deformation. Similarly, in \cite{scharff2018color}, bending deformation is detected through changes in colour ratios, observed by compact colour sensors.  While these approaches are foundational and important to addressing the deformation detection problem, they often come with certain drawbacks such as their dependency on manual feature selection and engineering, and their sensitivity to noise. 
Deep learning-based approaches, on the other hand, leverage the power of neural networks to automatically learn and extract features relevant to deformation. In \cite{zhao2020line}, Zhao et al., present an approach for the deformation prediction method using a deep learning model during the numerical control machining process. In \cite{tabernik2020segmentation}, introduce a segmentation-based methodology for automating the detection of surface anomalies, including deformations. Their strategy leveraged deep learning for segmentation and a decision network, enabling precise identification and analysis of irregularities on surfaces.  

Unsupervised Domain Adaptation (UDA) aims to leverage labelled data in one or more source domains to perform well on the unlabeled target domain, despite differences in the data distribution between these domains \cite{ben2010theory, fayyad2023out}. This learning paradigm is crucial for various computer vision tasks where acquiring labelled data is limited or highly costly. UDA can be classified into discrepancy-based \cite{cheng2024deep, jiang2023unsupervised, xiang2023discriminative, feng2023dda} reconstruction-based \cite{behrendt2023guided}, and adversarial-based approaches \cite{ran2024gradient, liu2024multi, wei2024self, cai2024symmetric}. Among these, adversarial methods have gained prominence due to their effectiveness in minimizing the domain discrepancy by framing the adaptation as a two-player game, where one network aims to distinguish between the source and target domains, while another tries to make the domains indistinguishable \cite{ganin2015unsupervised, tzeng2017adversarial}. This technique has been further refined by recent advancements such as the gradient-aligned domain adversarial network \cite{ran2024gradient}, which introduces novel mechanisms for enhancing the alignment of domain features, and the multi-level joint distribution alignment-based domain adaptation \cite{liu2024multi}, demonstrating the potential of adversarial methods in bridging domain gaps more effectively.

The utilization of Generative Adversarial Networks (GANs) in adversarial domain adaptation showcases significant advancements in domain transfer tasks. GANs, through their generator and discriminator networks, have been used for domain adaptation to either generate synthetic data that is indistinguishable from the target domain or learn a transformation from the source to the target domain. Notable implementations include the Domain-Adversarial Neural Network (DANN) \cite{ganin2015unsupervised}, which introduces a gradient reversal layer to fool the domain classifier, effectively making the feature extractor domain-invariant, and CycleGAN \cite{zhu2017unpaired}, which has been used for image-to-image translation tasks, demonstrating the capability to adapt domains without paired examples. In recent studies, Yao et al. introduced a Shared Wasserstein adversarial learning approach \cite{yao2024shared}, while Ge and Sadhu \cite{ge2024domain} utilized physics-informed learning and self-attention mechanisms to enhance GAN performance. Further enriching the domain, Wei et al. \cite{wei2024self} proposed a self-training method via weight transmission between generators, and Cai et al. \cite{cai2024symmetric} explored the potential of symmetric consistency with cross-domain mixup. These contributions collectively illustrate the methodological advancements GANs bring to UDA, driving the field towards more effective adaptation techniques.

\section{Methodology}

\subsection{Data Generation}
Blender is an open-source computer graphics, rendering, and simulation environment with a variety of tools that make it ideal for the creation of synthetic visual data of deformed and non-deformed objects. BlendTorch \cite{blendtorch_icpr2020_cheind} was used to help facilitate the data generation process by integrating the Python scripting environment within Blender into an integrated development environment. The object selected to be deformed was a pop can as the metallic surface features provide an interesting classification challenge and ease of physical dataset collection. For the synthetic dataset, a computer-aided design (CAD) file was imported and using expert information deformations were applied. A lattice deformation linked to the pop can was used to apply larger deformations to the structure while a hard \textit{Stucci} displace modifier was used on the surface of the can. These various deformations were all assigned shape keys enabling a Python script to create various procedural interpolations of these deformations. In both the deformed and non-deformed classes, shape keys were used to open the tab and seal of the cans. The material shading was set to mimic the metallic look of real-world pop cans and the outer label was wrapped around the object using a UV map. The scene in Blender was lit using randomized High Dynamic Range Images (HDRIs) given a random rotation to provide varietal realistic lighting. Camera rays that did not hit the object were set to black pixels giving the generated image a black background. Cameras were placed stochastically according to Table \ref{DG_tab1} and 6000 images were generated; 3000 deformed, 3000 non-deformed. The images were rendered as 512x512 RGB.

\begin{table}[htbp]
\centering
\caption{Soda Can Camera position parameters in Blender}\label{DG_tab1}
\begin{tabular}{@{}ccccc@{}}
\multicolumn{1}{l}{\multirow{2}{*}{}}
& \multicolumn{4}{c}{Camera} \\ \cmidrule(l){2-5} 
\multicolumn{1}{l}{} & No. 1 & No. 2 & No.3 & No.4 \\ \midrule
$\theta$ & $[20^{\circ},70^{\circ}]$ & $[110^{\circ},160^{\circ}]$ & $[200^{\circ},250^{\circ}]$ & $[290^{\circ},340^{\circ}]$ \\ \midrule
$\phi$ & \multicolumn{4}{c}{$[50^{\circ},70^{\circ}]$ for all cameras} \\ \midrule
$r$ & \multicolumn{4}{c}{$[0.3m-0.45m]$ for all cameras} \\ \bottomrule
\end{tabular}
\end{table}

\subsection{CycleGAN}
CycleGAN \cite{zhu2020unpaired}, is an unpaired image translation network that can be used for domain adaptation. This network has been shown to be effective in image-to-image translation, as well as neural style transfer. This architecture will be used as a benchmark to compare results against. The architecture of CycleGAN consists of two generators $G, F$ in competition with their respective discriminators $D_X, D_Y$. The generators are tasked with translating an image $I_X, I_Y$ into the other respective domain $I_{X \rightarrow Y}$, $I_{Y \rightarrow X}$ while ensuring enough data from the original image is present for the other generator to reconstruct the image $I'_X$, $I'_Y$.  

The architecture of the generator has 3 parts, an encoder, residual blocks, and a decoder. The encoder section consists of expanding convolutional layers increasing the channels from 3 to 64, to 128, and finally 256. A kernel size of 4 with a stride of 2 and a padding of 1 was used. The residual block portion consisted of 9 residual block layers with 256 channels. The decoder section is the exact inverse of the encoder architecture with decreasing channel sizes from 256 to 128, to 64 and finally 3. The final output from the generator is an image.

The discriminator consists of 6 convolutional layers, a linear layer, and then a sigmoid-activated 1 hot output. The convolutional layers first increase in channel size from 3 to 64, to 128, to 256, to 512 then finally decrease to 16. The single linear layer goes from 3136 to just 1. A sigmoid activation function is applied to the output to generate a probability the input image is generated or not.

\subsection{Content-Aware Sim-to-real Network (CASNet)}
\begin{figure*}
    \centering
    \includegraphics[width=0.98\textwidth]{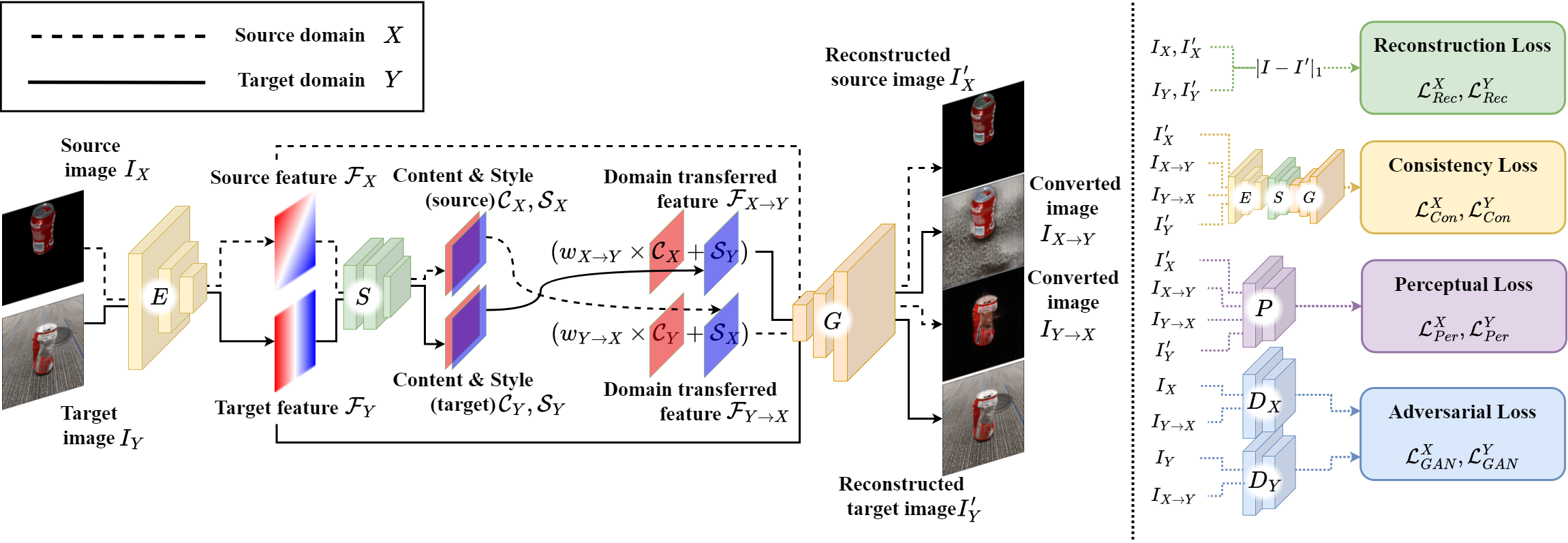}
    \caption{CASNet Architecture}
    \label{DRANet_Architecture}
\end{figure*}

The proposed architecture CASNet for domain adaptive deformation detection is based on the architecture of DRANet \cite{DRANet_CVPR2021}. DRANet is a disentangling representation and adaptation network which functions by taking two or more image datasets and gaining an implicit understanding of the content and style. The styles can then be swapped to generate new images retaining the same content but with stylistic elements of the other dataset. This concept will be used to convert the synthetic can dataset to mimic the real-world dataset. The architecture of CASNet is shown in Figure \ref{DRANet_Architecture}. CASNet uses an encoder $E$, a feature separator $S$, a generator $G$, two discriminators for source and target $D_X$, $D_Y$, and a perceptual network $P$.
The workflow through the network is to take an image from both domains $I_X$, $I_Y$, and use an encoder to create a feature space that contains the essential information from the images. The feature space then passes through the separator where the content and style are extracted and the styles are swapped. The generator then takes the content and style from the separator and decodes it to form an image. The discriminator analyzes the output of the converted images from the generator $I_{X\rightarrow Y}$, $I_{Y\rightarrow X}$ and compares them to the source images $I_X$, $I_Y$. There is a discriminator for each domain. The synthetic discriminator $D_X$ analyzes images converted from the physical domain and original synthetic images. The physical discriminator $D_Y$ analyzes images converted from the synthetic domain and the original synthetic images. These networks work against the generator to encourage more convincing generated images.

The encoder $E$ consists of a convolutional layer increasing the channels to 32, followed by 3 residual block layers. Finally, a last convolutional layer is used and expands the channels from 32 to 64. The task of the encoder is to take an input image and capture its features.
The separator $S$ consists of three convolutional layers with 64 channels. The purpose of the separator is to split the features from the encoder into content and style feature layers.
The generator $G$ consists of a transpose convolutional layer from 64 to 32, 3 residual block layers and a last transpose convolution layer from 32 to 3. A $\tanh$ activation function is used on the output. The purpose of the generator is to take content and style feature layers and generate an image. 
All layers use ReLU as the activation function and spectral normalization \cite{DBLP:journals/corr/abs-1802-05957spectral}.
The discriminator takes in images ($I_X, I_{Y\rightarrow X}$ for $D_X$, $I_Y, I{X\rightarrow Y}$ for $D_Y$) and analyzes if the given image is generated or real. The architecture for the patchGAN discriminator was convolution layers from 3 to 64, to 128, to 256, to 512 and finally down to 1 channel. The architecture uses LeakyReLU as an activation function with a negative slope of 0.2.
The perceptual network $P$ is a pre-trained version of VGG-19 on ImageNet1k. The role of the perceptual network is to analyze the images for content and style and help train the separator network. The perceptual network inputs the reconstructed $I'_X$, $I'_Y$ and converted images $I_{X\rightarrow Y}$, $I_{Y\rightarrow X}$ and creates feature maps from several middle layers. The network finds perceptual features to create constraints on the content and style of the images to train the separator network in an unsupervised manner.

Style transfer can struggle with complex scenes and features. This is particularly evident when the images have different scene structures and object compositions \cite{DRANet_CVPR2021}. This is particularly evident in trying to adapt the synthetic scene to the real-world scene. This manifests as \textit{ghosting} in this particular task. The two datasets seem to mix with content from another image appearing in the generated image. Content-Adaptive Domain Transfer (CADT) aims to solve this by searching the target features for content components most similar to the source features and then reflecting style information from more suitable target features. This is achieved by using a content similarity matrix $\mathcal{H}_{row}$.

\begin{equation}
    \label{eq:Hrow}
    \begin{split}
       \mathcal{H}_{row} = \sigma_{row} (\mathcal{C}_X \cdot \mathcal{C}_Y^\top) = \left[ \begin{array}{ccc}
       \mathcal{C}_{11}  & \cdots & \mathcal{C}_{1b} \\
        \vdots & \ddots & \vdots \\
        \mathcal{C}_{b1} & \cdots & \mathcal{C}_{bb}\\
    \end{array}\right] ,\\
    \text{where}\ \mathcal{C}_X, \mathcal{C}_Y \in \mathbb{R}^{B\times N} 
    \end{split}
\end{equation}

$\sigma_{row}$ is a softmax in the row dimension, the size of content factor $\mathcal{C}_X$ is defined by batch size $B$ and feature dimension $N$. This matrix is used to create the content-adaptive style feature described in the following equation:

\begin{equation}
    \hat{\mathcal{S}_Y} = \mathcal{H}_{row}\mathcal{S}_Y, \quad \text{where}\ \mathcal{S}_Y \in \mathbb{R}^{B\times N}
\end{equation}

Content-adaptive style transfer can be applied in the opposite direction by using the column direction and the following equation:

\begin{equation}
    \mathcal{H}_{col} = (\sigma_{col}(\mathcal{C}_X \cdot \mathcal{C}_Y^\top ))^\top
\end{equation}

The results using CADT stylistically are much better, with \textit{ghosting} artifacts removed and are essential for proper data generation for deformation detection purposes.

\subsection{VGG-16 Classifier}
A CNN-based classifier VGG-16 \cite{VGG-16} is used to analyze the images and classify the objects as being deformed or non-deformed. A pre-trained version of the network on ImageNet1k \cite{deng2009imagenet} was used with modifications after convolution layers. After the convolution layers, an adaptive average pooling layer is used to fix the output size. This is then followed by 3 fully connected layers with ReLU activation and a dropout layer. The 3 fully connected layers go to a one-hot output with a sigmoid activation function.

\section{Training}
\subsection{CycleGAN}
CycleGAN was trained on the synthetic and real datasets for 200 steps with a learning rate of 0.001 and a batch size of 32 using the Adam optimizer. After each batch, the network parameters were updated. To reduce oscillation spectral normalization was used in the discriminator. The generator was trained for 2 cycles for each training step and the discriminator was trained once. Longer training times were investigated but yielded similar results.

\subsection{CASNet}
In the original implementation of DRANet, the classifier is trained at the same time as the GAN network. In this implementation, the training of these networks was separated. The GAN network was trained first, then all the images were generated and saved. The classifier was then trained on the saved converted images. CASNet was trained for 2000 steps using the Adam optimizer with a learning rate of 0.001 and a batch size of 2.

\subsection{VGG-16 Classifier}

While training the classifier, all convolution layers except the final layer were frozen to preserve learned features from its previous training on ImageNet-1k. This pre-training allows the model to learn the general features of images, which can be transferred to classifying the state of cans. The VGG-16 classifier was trained on the generated images for 15 epochs using the SGD optimizer. The learning rate was set to 0.005 and a batch size of 32 was used.

\section{Results and Discussion}

\begin{figure*}[htbp]
\centering
     \begin{subfigure}[b]{0.32\textwidth}
         \centering
         \includegraphics[width=0.98\textwidth]{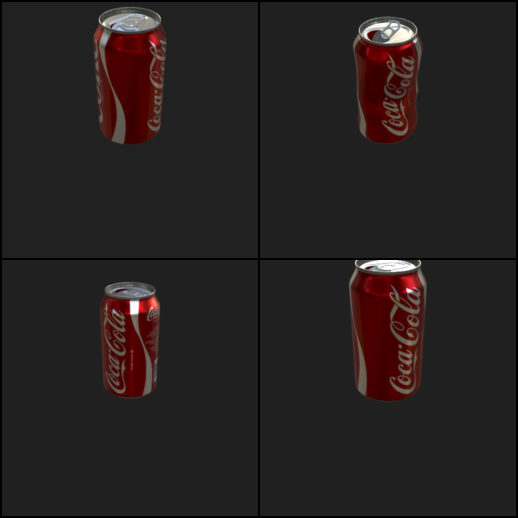}
         \caption{Black Background CycleGAN Input\\
         \hspace{10mm}}
         \label{DA_CycleGAN_in}
     \end{subfigure}
     \begin{subfigure}[b]{0.32\textwidth}
         \centering
         \includegraphics[width=0.98\textwidth]{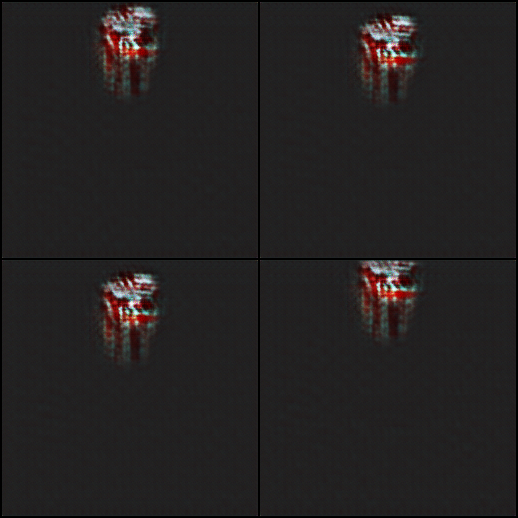}
         \caption{Black Background CycleGAN Reconstruction}
         \label{DA_CycleGAN_recon}
     \end{subfigure}
     \begin{subfigure}[b]{0.32\textwidth}
         \centering
         \includegraphics[width=0.98\textwidth]{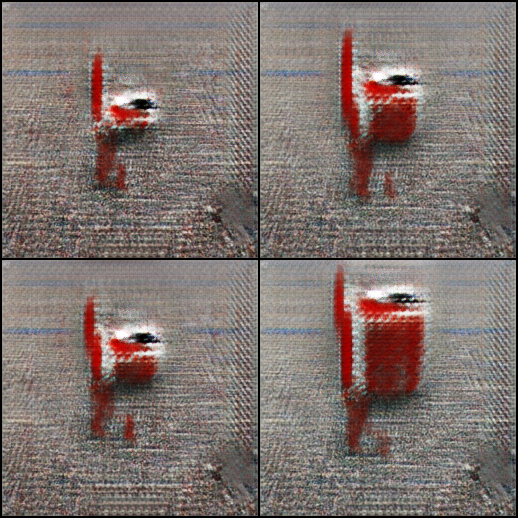}
         \caption{Black Background CycleGAN Converted \\ \hspace{0.5cm}}
         \label{DA_CycleGAN_convert}
     \end{subfigure}
    \caption{CycleGAN Image Analysis}
    \label{DA_CycleGAN}
\end{figure*}

\begin{figure*}[htbp]
\centering
     \begin{subfigure}[b]{0.32\textwidth}
         \centering
         \includegraphics[width=0.98\textwidth]{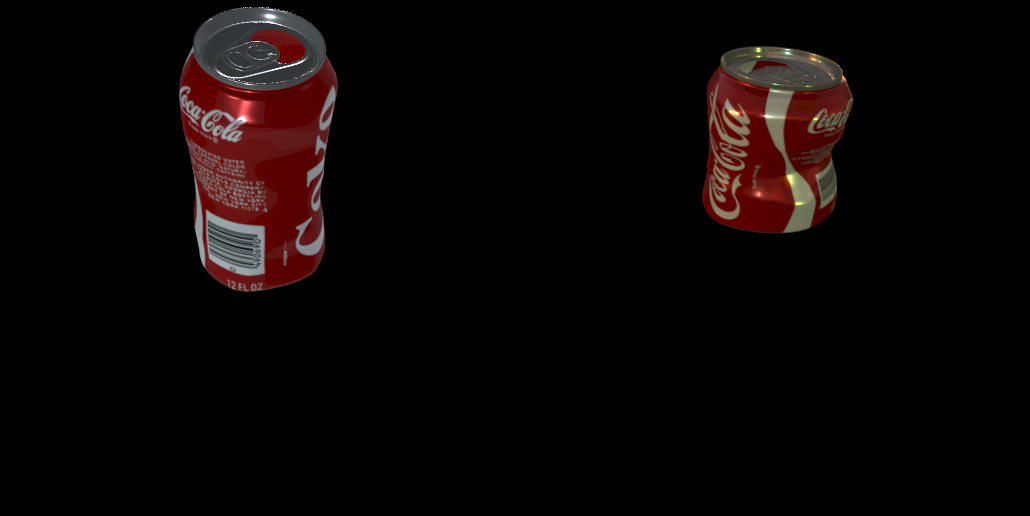}
         \caption{Input Images from Black Background Dataset}
         \label{dranet_SC_input}
     \end{subfigure}
     \begin{subfigure}[b]{0.32\textwidth}
         \centering
         \includegraphics[width=0.98\textwidth]{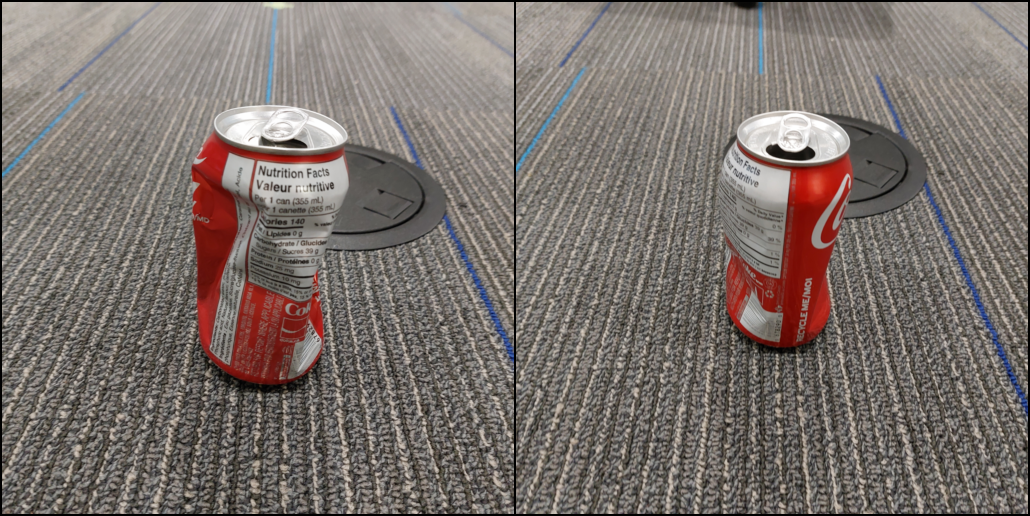}
         \caption{Input Images from Real-World Dataset \\ \hspace{0.5cm} }
         \label{dranet_PC_input}
     \end{subfigure}
     \begin{subfigure}[b]{0.32\textwidth}
         \centering
         \includegraphics[width=0.98\textwidth]{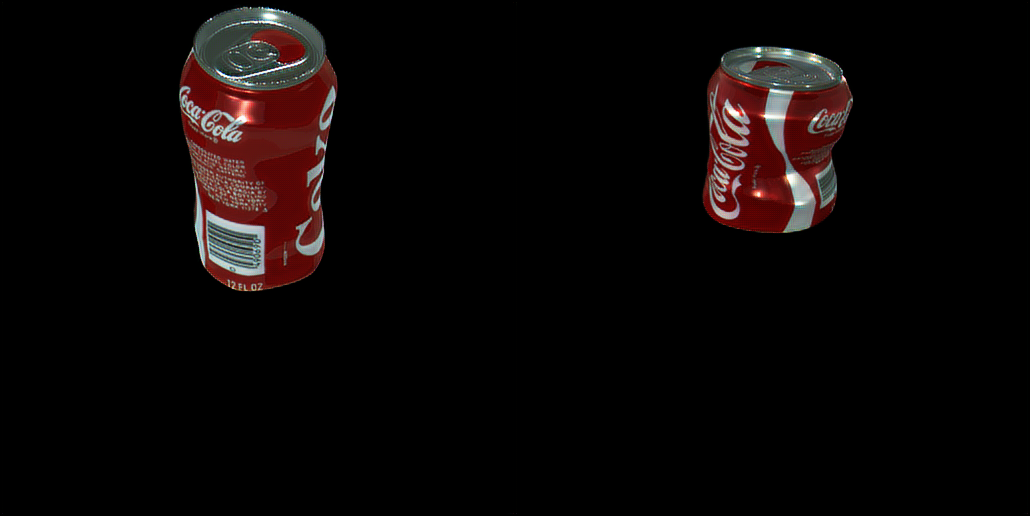}
         \caption{Reconstructed Images from Black Background Dataset}
         \label{dranet_SC_recon}
     \end{subfigure}
     \begin{subfigure}[b]{0.32\textwidth}
         \centering
         \includegraphics[width=0.98\textwidth]{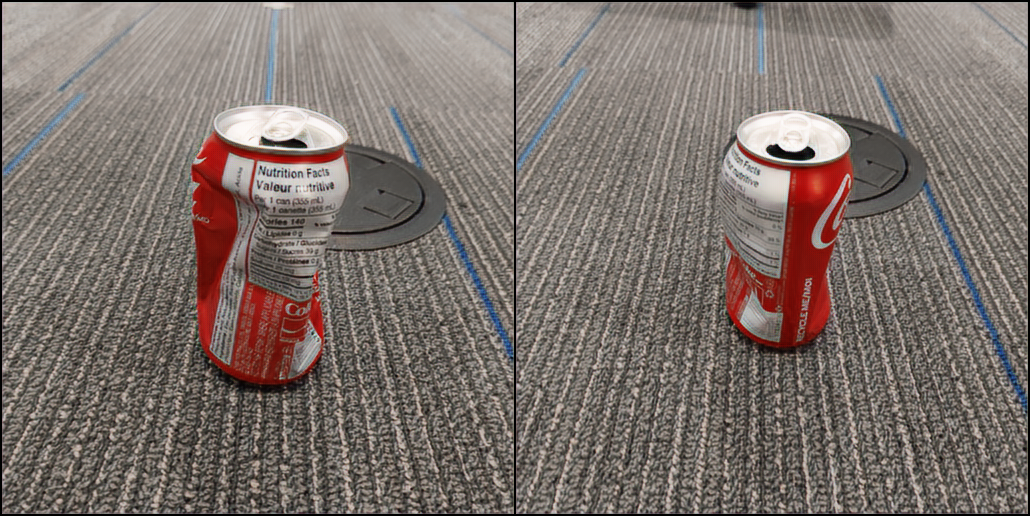}
         \caption{Reconstructed Images from Real-World Dataset}
         \label{dranet_PC_recon}
     \end{subfigure}
     \begin{subfigure}[b]{0.32\textwidth}
         \centering
         \includegraphics[width=0.98\textwidth]{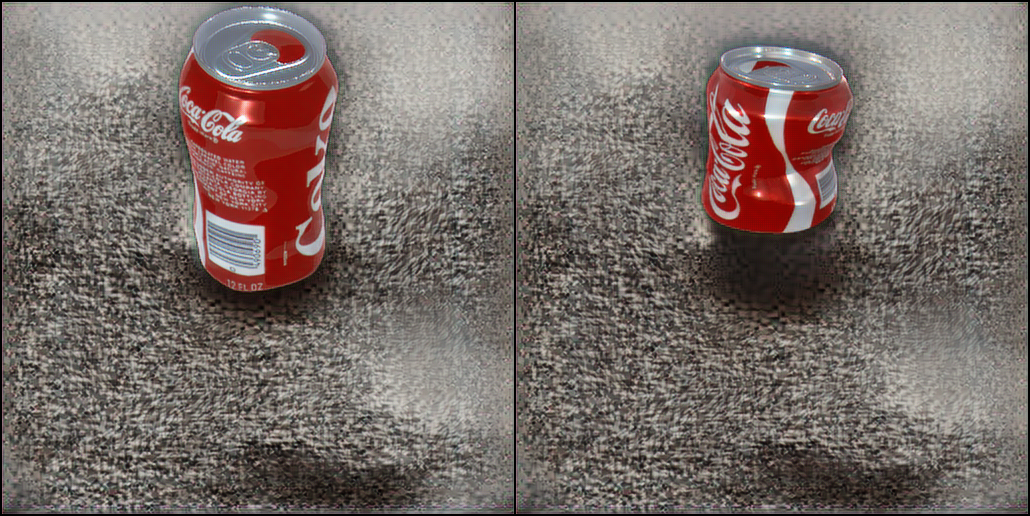}
         \caption{Converted Images from Black Background Dataset}
         \label{dranet_SC_convert}
     \end{subfigure}
     \begin{subfigure}[b]{0.32\textwidth}
         \centering
         \includegraphics[width=0.98\textwidth]{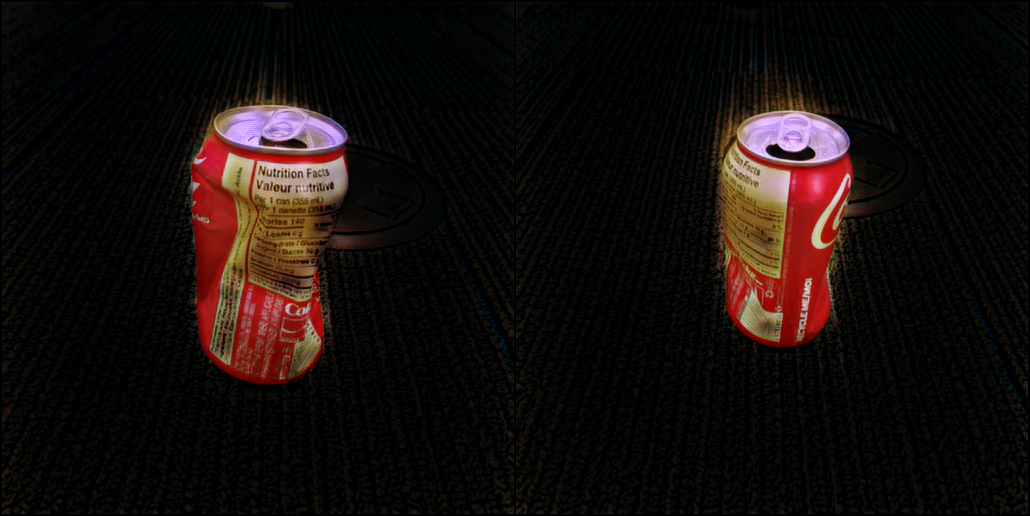}
         \caption{Converted Images from Real-World Dataset}
         \label{dranet_PC_convert}
     \end{subfigure}
    \caption{Domain Adaptation Results from CASNet}
    \label{DRANet_imgs}
\end{figure*}

The images resulting from training CycleGAN and CASNet are shown in Figures \ref{DA_CycleGAN} and \ref{DRANet_imgs} respectively. The generated data from CycleGAN is shown in Figure \ref{DA_CycleGAN_convert} and shows poor conversion to the real-world domain. The background of the image seems to have converted quite well in the image but there is a significant degradation in the appearance of the can. The can has become heavily distorted and seems to also have a limited variation. The reconstructed image shown in Figure \ref{DA_CycleGAN_recon} also shows poor reconstruction showing that CycleGAN struggles significantly with this task. The results produced by CycleGAN proved unsatisfactory for image translation between the synthetic and real-world datasets and classification was not performed on the data produced by this architecture.

Using the input images from the synthetic and real-world datasets in Figures \ref{dranet_SC_input} and \ref{dranet_PC_input}, CASNet was effectively able to discover the content and styles of the images. The converted images in Figures \ref{dranet_SC_convert} and \ref{dranet_PC_convert} have excellent results. While the background of the synthetic cans isn't very high quality, the content transferred is excellent. The background of the converted synthetic images is a decent attempt at transferring the surface the cans are placed on from the real-world dataset. The black background from the synthetic dataset is applied to the background of the real-world dataset well and shows excellent style transfer. Overall the converted image quality from CASNet is excellent and shows a great understanding of the content and style of the images. The reconstructed images in Figures \ref{dranet_SC_recon} and \ref{dranet_PC_recon} are very similar to the originals which show that CASNet is performing well and is suitable for further deformation classification training.

Only the converted images from synthetic to real are used for creating the generated dataset for further classification. However, in different applications or circumstances, the real-world images converted to synthetic could be used as well.

\begin{figure}[htbp]
    \centering
    \includegraphics[width=0.48\textwidth]{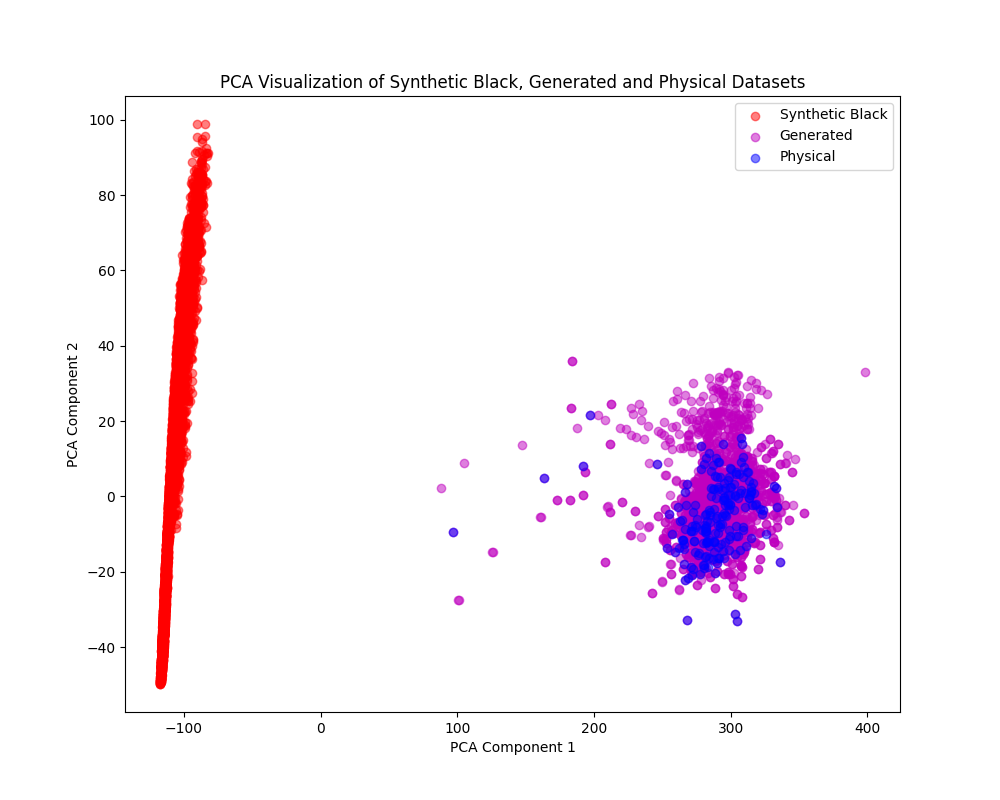}       
    \caption{CASNet Generated Images PCA Visualizations}
    \label{DRANet_PCA}
\end{figure}

The PCA visualizations shown in Figure \ref{DRANet_PCA} show that CASNet was effective at capturing the real-world dataset and was successful in its domain transfer task. The generated images capture the physical distribution well and validate CASNet's effectiveness in domain transfer.

\begin{figure}[htbp]
\centering
    \begin{subfigure}[b]{0.23\textwidth}
         \centering
         \includegraphics[width=0.98\textwidth]{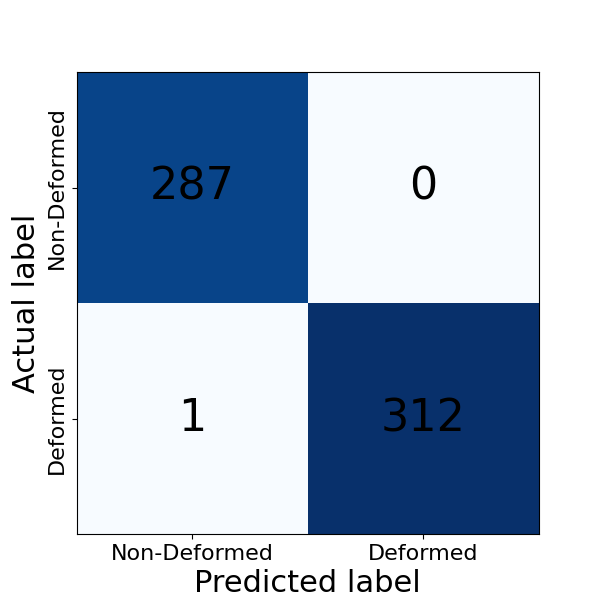}
         \caption{Black Background - Synthetic dataset}
         \label{confusion_black_1}
     \end{subfigure}
     \begin{subfigure}[b]{0.23\textwidth}
         \centering
         \includegraphics[width=0.98\textwidth]{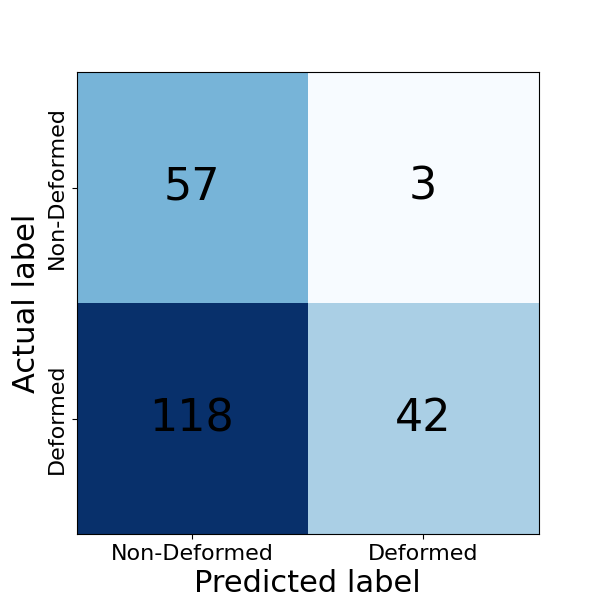}
         \caption{Black Background - Sim-to-Real}
         \label{confusion_black_2}
     \end{subfigure}
     \begin{subfigure}[b]{0.23\textwidth}
         \centering
         \includegraphics[width=0.98\textwidth]{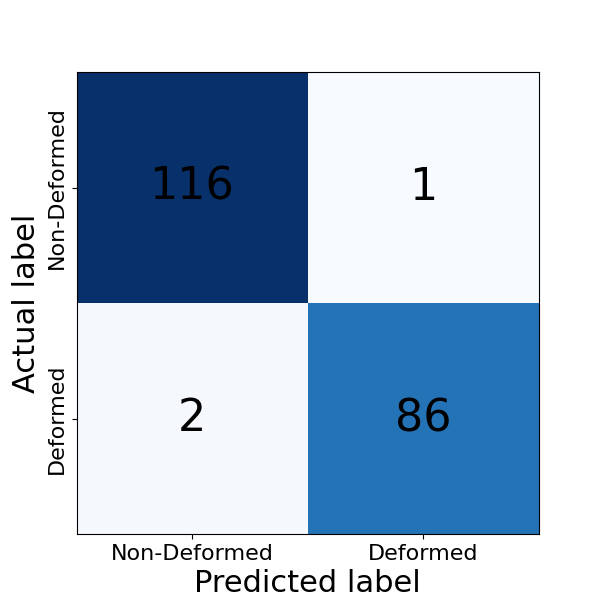}
         \caption{Generated Dataset - Synthetic dataset}
         \label{confusion_1}
     \end{subfigure}
     \begin{subfigure}[b]{0.23\textwidth}
         \centering
         \includegraphics[width=0.98\textwidth]{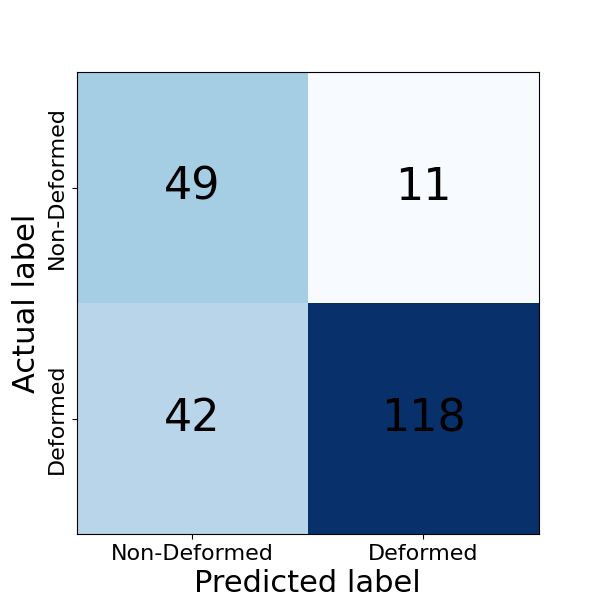}
         \caption{Generated Dataset - Sim-to-Real}
         \label{confusion_2}
     \end{subfigure}
    \caption{Confusion matrices for Comparing CASNet Performance}
    \label{confusion}
\end{figure}

The classification performance metrics for VGG-16 are shown in Table \ref{Dda_tab1} for the black background and generated datasets. The confusion matrices are shown in Figure \ref{confusion} and provide a breakdown of the results for the deformed and non-deformed classes. For the \textit{Synthetic dataset} case VGG-16 was trained and evaluated on synthetic or generated data only where the dataset was split into training and testing datasets. The \textit{Sim-to-real} case is where VGG-16 was trained on the generated converted data and then evaluated on the real-world dataset. The network achieved excellent metrics evaluating on the synthetic data as shown in Figures \ref{confusion_black_1} and \ref{confusion_black_1}. However, the performance dropped when the network was evaluated on real-world data as shown in Figures \ref{confusion_black_2} and \ref{confusion_2}. The metrics in Table \ref{Dda_tab1} show when VGG-16 was trained on the data generated by CASNet there was a significant increase in accuracy, F1-score and precision compared to training on the black background data. The domain transfer provided by CASNet results in a considerable increase in sim-to-real performance. 

\begin{table}[pbht]
\centering
\caption{CASNet Comparative Performance to Best Data Augmentation}\label{Dda_tab1}
\begin{tabular}{lcc|cc}
\multirow{2}{*}{Metrics} & \multicolumn{2}{c|}{Synthetic dataset} & \multicolumn{2}{c}{Sim-to-real} \\ \cline{2-5}
 & Black & Generated & Black & Generated \\ \hline
{$Accuracy$\quad} & \textbf{0.998} & 0.985 & 0.450 & \textbf{0.759} \\ \hline
{$F_1$\quad} & \textbf{0.998} & 0.987 & 0.485 & \textbf{0.649} \\ \hline
{$Recall$\quad} & \textbf{1.0} & 0.991 & \textbf{0.950} & 0.817 \\ \hline
{$Precision$\quad} & \textbf{0.997} & 0.983 & 0.326 & \textbf{0.538} \\ \hline
\end{tabular}
\end{table}

Comparing the generated images from CycleGAN and CASNet shows that the best performance is from CASNet. CycleGAN produced images that copied the style of the background quite well but had a very negative effect on the content of the soda can. The cans in the images produced very loosely resemble their original look and the reconstruction doesn't match the original image. CASNet produced images where the content is clear and matches the content of the source images. The backgrounds in the generated images are a decent attempt at recreating the target background but have room for improvement. A possible avenue for improving the background style transfer for CASNet could be replacing the Gram style loss with Sliced Wasserstein Discrepancy (SWD) \cite{Apple_SWD_CVPR}. Experimentation was done with SWD style loss but results had issues where the content sometimes failed to transfer. Overall the style transfer was improved with this loss but additional work is required to ensure proper content transfer.

The sim-to-real results from training using the dataset generated by CASNet show considerable improvements in F1-score and precision as well as a slight improvement in accuracy in comparison to the original synthetic data. This can be attributed to the generated data distribution better matching the physical dataset in comparison to the black background dataset. Additional improvement to these classification metrics could be possible with a different network architecture or by changing the datasets to include depth information.

\section{Conclusion}
Geometrical quality assurance is an important aspect of industrial processes. The method shown in this work can be used to bring modern machine learning techniques to quality assurance processes that can be expensive and difficult. This is particularly useful in cases where the object of quality control is difficult to create a dataset, e.g., the object is fragile, difficult to handle, expensive, and visual methods are required. The novelty of this work is the use of CASNet for sim-to-real domain adaptation for the use of deformation classification. In this work, CASNet was shown to be able to effectively adapt synthetic datasets from Blender to real-world distributions allowing for better, more realistic data and improved classification metrics.  These adapted datasets have a dramatic improvement to deformation classification metrics and reduce the sim-to-real gap present in synthetic datasets. A strength of this method is that minimal unlabelled real-world data is required which reduces time and cost. Our ongoing research aims to improve the style transfer of CASNet using sliced Wasserstein discrepancy. In future work, we plan to incorporate depth information into the datasets as it represents a promising avenue for enhancing model performance. Depth data provides a critical spatial understanding that can significantly improve the accuracy and robustness of the model, especially for tasks requiring precise object localization and scene comprehension. Additionally, exploring different network architectures such as vision transformers could yield better performance.

{
\bibliographystyle{IEEEtran}
\bibliography{ref}
}

\end{document}